%% file: _main.tex
\input{_constants}
\arxiv 

\pdfoutput=1
\documentclass[10pt,twocolumn,letterpaper]{article}
\input{cvpr_header}

\unless\ifarxiv \myexternaldocument{_supplementary} \fi

\begin{document}
\title{\paperTitle}
\author{\authorBlock}
\maketitle

\input{00_abstract}
\input{01_intro}

\input{02_solution}
\input{03_experiments}

{\small
\bibliographystyle{ieeenat_fullname}
\bibliography{11_references}
}


\end{document}


\title{\paperTitle}
\author{\authorBlock}
\maketitlesupplementary

\appendix
\input{12_appendix}

{\small
\bibliographystyle{ieeenat_fullname}
\bibliography{11_references}
}

%% file: _constants.tex
\def\paperID{0000}
\def\confName{CVPR}
\def\confYear{2024}

\def\paperTitle{Hydra-MDP: End-to-end Multimodal Planning with Multi-target Hydra-Distillation}

\def\authorBlock{
    Zhenxin Li\textsuperscript{1, 2} \quad
    Kailin Li\textsuperscript{3} \quad
    Shihao Wang\textsuperscript{1, 4} \quad 
    Shiyi Lan\textsuperscript{1} \quad
    Zhiding Yu\textsuperscript{1}\quad
    Yishen Ji\textsuperscript{5} \\
    Zhiqi Li\textsuperscript{5} \quad
    Ziyue Zhu\textsuperscript{6} \quad
    Jan Kautz\textsuperscript{1} \quad
    Zuxuan Wu\textsuperscript{2} \quad
    Yu-Gang Jiang\textsuperscript{2} \quad
    Jose M. Alvarez\textsuperscript{1} \\
  {$^1$NVIDIA} \quad {$^2$Fudan University} \quad {$^3$East China Normal University} \\ {$^4$Beijing Institute of Technology} \quad {$^5$Nanjing University} \quad {$^6$Nankai University}
}

\newif\ifreview 
\newif\ifarxiv \newcommand{\arxiv}{\arxivtrue}
\newif\ifcamera 
\newif\ifrebuttal 

%% file: cvpr_header.tex
\ifreview \usepackage[review]{cvpr} \fi
\ifarxiv \usepackage[pagenumbers]{cvpr} \fi
\ifrebuttal \usepackage[rebuttal]{cvpr} \fi
\ifcamera \usepackage{cvpr} \fi

\input{_macros}  

\usepackage{xr-hyper}

\makeatletter
\newcommand*{\addFileDependency}[1]{
  \typeout{(#1)}
  \@addtofilelist{#1}
  \IfFileExists{#1}{}{\typeout{No file #1.}}
}

\makeatother
\newcommand*{\myexternaldocument}[1]{
    \externaldocument{#1}
    \addFileDependency{#1.tex}
    \addFileDependency{#1.aux}
}

\definecolor{cvprblue}{rgb}{0.21,0.49,0.74}
\usepackage[pagebackref,breaklinks,colorlinks,citecolor=cvprblue]{hyperref}
\usepackage[capitalize]{cleveref}
\crefname{section}{Sec.}{Secs.}
\crefname{table}{Table}{Tables}
\crefname{figure}{Fig.}{Figs.}

\ifarxiv \crefname{appendix}{App.}{Apps.}
\else \crefname{appendix}{Suppl.}{Suppls.} \fi

%% file: _macros.tex
\usepackage{graphicx}	
\usepackage{amsmath}	
\usepackage{amssymb}	
\usepackage{booktabs}
\usepackage{times}
\usepackage{microtype}
\usepackage{epsfig}
\usepackage[table,xcdraw,dvipsnames]{xcolor}
\usepackage{caption}
\usepackage{float}
\usepackage{placeins}
\usepackage{color, colortbl}
\usepackage{stfloats}
\usepackage{enumitem}
\usepackage{tabularx}
\usepackage{xstring}
\usepackage{multirow}
\usepackage{xspace}
\usepackage{url}
\usepackage{subcaption}
\usepackage{xcolor}
\usepackage[hang,flushmargin]{footmisc}

\ifcamera \usepackage[accsupp]{axessibility} \fi

% ...

\ifarxiv  \fi

\newcommand{\hydra}{Hydra-MDP}

\newcommand{\R}[1]{{%
    \textbf{%
        \ifstrequal{#1}{1}{\textcolor{red}{R#1}}{%
        \ifstrequal{#1}{2}{\textcolor{blue}{R#1}}{%
        \ifstrequal{#1}{3}{\textcolor{magenta}{R#1}}{%
        \ifstrequal{#1}{4}{\textcolor{teal}{R#1}}{%
                           \textcolor{cyan}{R#1}%
        }}}}%
    }%
}}

%% file: 00_abstract.tex
\begin{abstract}
We propose Hydra-MDP, a novel paradigm employing multiple teachers in a teacher-student model. 
This approach uses knowledge distillation from both human and rule-based teachers to train the student model, which features a multi-head decoder to learn diverse trajectory candidates tailored to various evaluation metrics. 
With the knowledge of rule-based teachers, Hydra-MDP learns how the environment influences the planning in an end-to-end manner instead of resorting to non-differentiable post-processing. 
This method achieves the $1^{st}$ place in the Navsim challenge, demonstrating significant improvements in generalization across diverse driving environments and conditions. More details by visiting \url{https://github.com/NVlabs/Hydra-MDP}.

\end{abstract}

%% file: 01_intro.tex
\vspace{-0.1in}
\section{Introduction}
\label{sec:intro}

End-to-end autonomous driving, which involves learning a neural planner with raw sensor inputs, is considered a promising direction to achieve full autonomy.
Despite the promising progress in this field~\cite{hu2023planning, jiang2023vad}, recent studies~\cite{dauner2023parting, li2023ego, chen2024vadv2} have exposed multiple vulnerabilities and limitations of imitation learning (IL) methods,
particularly the inherent issues in open-loop evaluation, such as the dysfunctional metrics and implicit biases~\cite{li2023ego,dauner2023parting}. This is critical as it fails to guarantee safety, efficiency, comfort, and compliance with traffic rules. To address this main limitation, several works have proposed incorporating closed-loop metrics, which more effectively evaluate end-to-end autonomous driving by ensuring that the machine-learned planner meets essential criteria beyond merely mimicking human drivers.

Therefore, end-to-end planning is ideally a multi-target and multimodal task, where multi-target planning involves meeting various evaluation metrics from either open-loop and closed-loop settings. In this context, multimodal indicates the existence of multiple optimal solutions for each metric. 

Existing end-to-end approaches~\cite{chen2024vadv2, jiang2023vad, hu2023planning} often try to consider closed-loop evaluation via post-processing, which is not streamlined
and may result in the loss of additional information compared to a fully end-to-end pipeline. Meanwhile, rule-based planners~\cite{dauner2023parting, treiber2000congested} struggle with imperfect perception inputs. These imperfect inputs degrade the performance of rule-based planning under both closed-loop and open-loop metrics, as they rely on predicted perception instead of ground truth (GT) labels.

\input{figs/teaser}

To address the issues, we propose a novel end-to-end autonomous driving framework called Hydra-MDP (Multimodal Planning with Multi-target Hydra-distillation). Hydra-MDP is based on a novel teacher-student knowledge distillation (KD) architecture. The student model learns diverse trajectory candidates tailored to various evaluation metrics through KD from both human and rule-based teachers. We instantiate the multi-target Hydra-distillation with a multi-head decoder, thus effectively integrating the knowledge from specialized teachers. Hydra-MDP also features an extendable KD architecture, allowing for easy integration of additional teachers.

The student model uses environmental observations during training, while the teacher models use ground truth (GT) data. This setup allows the teacher models to generate better planning predictions, helping the student model to learn effectively. By training the student model with environmental observations, it becomes adept at handling realistic conditions where GT perception is not accessible during testing.

Our contributions are summarized as follows:

\begin{enumerate}
    \item We propose a universal framework of end-to-end multimodal planning via multi-target hydra-distillation, allowing the model to learn from both rule-based planners and human drivers in a scalable manner.
    \item Our approach achieves the state-of-the-art performance under the simulation-based evaluation metrics on Navsim.
\end{enumerate}

%% file: figs/teaser.tex
\begin{figure}
    \centering
        \includegraphics[width=\linewidth]{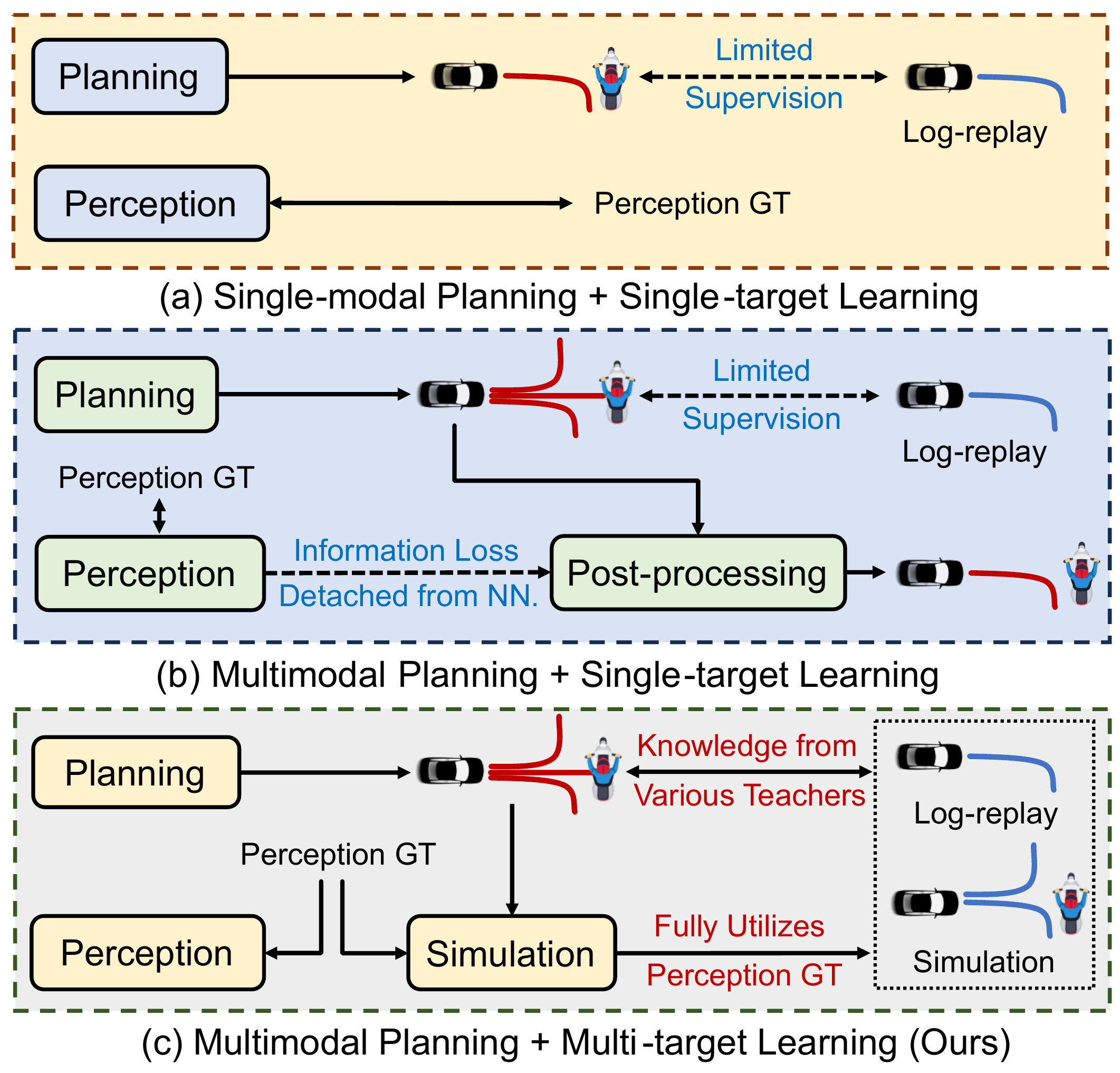}
    \vspace{-0.2in}

    \caption{\textbf{Comparison between End-to-end Planning Paradigms.}}
    \vspace{-0.2in}
\label{fig:teaser}
\end{figure}

%% file: 02_solution.tex
\newcommand{\pred}[0]{$\mathcal{S}^m_i$}
\newcommand{\gt}[0]{$\hat{\mathcal{S}}^m_i$}
\section{Solution}
\label{sec:solution}
\input{figs/arch}

\subsection{Preliminaries}
Let $O$ represent sensor observations,  $\hat{P}$ and $P$ denote ground truth and predicted perceptions (\eg 3D object detection, lane detection), $\hat{T}$ be the expert trajectory, and $T^*$ be the predicted trajectory. 
$\mathcal{L}_{im}$ represents the imitation loss.
We first introduce the two prevailing paradigms and our proposed paradigm (Fig.~\ref{fig:teaser}) in this section:

\noindent\textbf{A. Single-modal Planning + Single-target Learning.} In this paradigm~\cite{hu2023planning, jiang2023vad, li2023ego}, the planning network directly regresses the planned trajectory from the sensor observations. Ground truth perceptions can be used as auxiliary supervision but does not influence the planning output. Perception losses are not included in the formula for simplicity. The whole processing can be formulated as:
\begin{equation}
\mathcal{L}=\mathcal{L}_{im}(T^*, \hat{T}),
\vspace{-0.15cm}
\end{equation}
where $\mathcal{L}_{im}$ is usually an L2 loss.

\noindent\textbf{B. Multimodal Planning + Single-target Learning.} This approach~\cite{chen2024vadv2, biswas2024quad} predicts multiple trajectories $\{T_i\}_{i=1}^k$, whose similarities to the expert trajectory are computed:
\begin{equation}
    \mathcal{L}=\sum_i \mathcal{L}_{im}(T_i, \hat{T}), 
    \vspace{-0.15cm}
\end{equation}
where $\mathcal{L}_{im}$ can be KL-Divergence~\cite{chen2024vadv2} or the max-margin loss~\cite{biswas2024quad}.
Perception outputs $P$ are explicitly used to post-process suitable trajectories via a cost function $f(T_i, P)$. The trajectory with the lowest cost is selected:
\begin{equation}
T^*=\underset{T_i}{\arg \min } f(T_i, P),
\vspace{-0.15cm}
\end{equation}
which is a non-differentiable process based on imperfect perception $P$.

\noindent\textbf{C. Multimodal Planning + Multi-target Learning.}
We propose this paradigm to simultaneously predict various costs (e.g., collision cost, drivable area compliance cost) via a neural network $\Tilde{f}$.
This is performed in a teacher-student distillation manner, where the teacher has access to ground truth perception $\hat{P}$ but the student relies only on sensor observations $O$. 
This paradigm can be formulated as:
\begin{equation}
    \mathcal{L}=\sum_i \mathcal{L}_{im}(T_i, \hat{T}) + \mathcal{L}_{kd}(f(T_i, \hat{P}), \Tilde{f}(T_i, O)). 
    \label{eq:para_c}
    \vspace{-0.15cm}
\end{equation}
Here, we only consider one cost function $f$ for clarity. 
The trajectory with the lowest predicted cost is selected:
\begin{equation}
    T^*=\underset{T_i}{\arg \min } \Tilde{f}(T_i, O).
    \vspace{-0.15cm}
\end{equation}

We stress that this framework is not restricted by non-differentiable post-processing.
It can be easily scaled in an end-to-end fashion by involving more cost functions or leveraging imitation similarity in our implementation (Sec.~\ref{subsec:inference}).

\subsection{Overall Framework}
As shown in Fig.~\ref{fig:arch}, \hydra{} consists of two networks: a \textbf{Perception Network} and a \textbf{Trajectory Decoder}. 

\vspace{0.05in}\noindent\textbf{Perception Network.} Our perception network builds upon the official challenge baseline Transfuser~\cite{chitta2022transfuser, Contributors2024navsim}, which consists of an image backbone, a LiDAR backbone, and perception heads for 3D object detection and BEV segmentation.
Multiple transformer layers~\cite{vaswani2017attention} connect  features from stages of both backbones, extracting meaningful information from different modalities.
The final output of the perception network comprises environmental tokens $F_{env}$, which encode abundant semantic information derived from both images and LiDAR point clouds.

\vspace{0.05in}\noindent\textbf{Trajectory Decoder.} 
Following Vadv2~\cite{chen2024vadv2}, we construct a fixed planning vocabulary to discretize the continuous action space. 
To build the vocabulary, we first sample 700K trajectories randomly from the original nuPlan database~\cite{caesar2021nuplan}.
Each trajectory $T_i(i=1,...,k)$ consists of 40 timestamps of $(x,y,heading)$, corresponding to the desired 10Hz frequency and a 4-second future horizon in the challenge.
The planning vocabulary $\mathcal{V}_k$ is formed as K-means clustering centers of the 700K trajectories, where $k$ denotes the size of the vocabulary.
$\mathcal{V}_k$ is then embedded as $k$ latent queries with an MLP, 
sent into layers of transformer encoders~\cite{vaswani2017attention}, and added to the ego status $E$:
\begin{equation}
    \mathcal{V}'_k = Transformer(Q,K,V=Mlp(\mathcal{V}_k))+E.
    \vspace{-0.15cm}
\end{equation} 
To incorporate environmental clues in $F_{env}$, transformer decoders are leveraged:
\begin{equation}
    \mathcal{V}''_k = Transformer(Q=\mathcal{V}'_k, K,V=F_{env}).
    \vspace{-0.15cm}
\end{equation}
Using the log-replay trajectory $\hat{T}$, we implement a distance-based cross-entropy loss to imitate human drivers: \begin{equation}
    \mathcal{L}_{im}=-\sum_{i=1}^{k} y_i\log(\mathcal{S}^{im}_i),
    \vspace{-0.15cm}
\end{equation} where $\mathcal{S}^{im}_i$ is the $i$-th softmax score of $\mathcal{V}''_k$, and $y_i$ is the imitation target produced by L2 distances between log-replays and the vocabulary. Softmax is applied on L2 distances to produce a probability distribution:
\begin{equation}
    y_i=\frac{e^{-(\hat{T}-T_i)^2}}{\sum_{j=1}^{k} e^{-(\hat{T}-T_j)^2}}.
    \vspace{-0.15cm}
\end{equation}
The intuition behind this imitation target is to reward trajectory proposals that are close to human driving behaviors.

\subsection{Multi-target Hydra-Distillation}

Though the imitation target provides certain clues for the planner, it is insufficient for the model to associate the planning decision with the driving environment under the closed-loop setting, 
leading to failures such as collisions and leaving drivable areas~\cite{li2023ego}.
Therefore, to boost the closed-loop performance of our end-to-end planner, we propose Multi-target Hydra-Distillation, a learning strategy that aligns the planner with simulation-based metrics in this challenge.

The distillation process expands the learning target through two steps: 
(1) running offline simulations~\cite{dauner2023parting} of the planning vocabulary $\mathcal{V}_k$ for the entire training dataset; 
(2) introducing supervision from simulation scores for each trajectory in $\mathcal{V}_k$ during the training process.
For a given scenario, step 1 generates ground truth simulation scores $\{$\gt$|i=1,...,k\}_{m=1}^{|M|}$ for each metric $m\in M$ and the $i$-th trajectory, where $M$ represents the set of closed-loop metrics used in the challenge.
For score predictions, latent vectors $\mathcal{V}''_k$ are processed with a set of Hydra Prediction Heads, yielding predicted scores $\{$\pred$|i=1,...,k\}_{m=1}^{|M|}$. With a binary cross-entropy loss, we distill rule-based driving knowledge into the end-to-end planner:
\begin{equation}
\scalebox{0.9}{$\mathcal{L}_{kd}=-\sum_{m, i} \hat{\mathcal{S}}^m_i \log \mathcal{S}^m_i + (1-\hat{\mathcal{S}}^m_i) \log (1 - \mathcal{S}^m_i).$}
\vspace{-0.15cm}
\end{equation}
For a trajectory $T_i$, its distillation loss of each sub-score acts as a learned cost value in Eq.~\ref{eq:para_c}, measuring the violation of particular traffic rules associated with that metric.

\subsection{Inference and Post-processing}
\label{subsec:inference}
\subsubsection{Inference}
Given the predicted imitation scores $\{\mathcal{S}^{im}_i|i=1,...,k\}$ and metric sub-scores $\{$\pred$|i=1,...,k\}_{m=1}^{|M|}$, we calculate an assembled cost measuring the likelihood of each trajectory being selected in the given scenario as follows:
\begin{flalign}
\Tilde{f}(T_i, O) = &-(w_1\log{\mathcal{S}^{im}_i} + w_2\log{\mathcal{S}^{NC}_i} + w_3\log{\mathcal{S}^{DAC}_i} \nonumber &&\\
&+ w_4\log{(5\mathcal{S}^{TTC}_i}+2\mathcal{S}^C_i+5\mathcal{S}^{EP}_i)),
\end{flalign} where $\{w_{i}\}_{i=1}^4$ represent confidence weighting parameters to mitigate the imperfect fitting of different teachers. 
The optimal combination of weights is obtained via grid search, which typically fall within the following ranges: $0.01 \leq w_1 \leq 0.1, 0.1 \leq w_2, w_3 \leq 1, 1\leq w_4\leq 10$, indicating the necessity to prioritize rule-based costs over imitation.
Finally, the trajectory with the lowest overall cost is chosen.
\vspace{-0.3cm}
\subsubsection{Model Ensembling}
We present two model ensembling techniques: Mixture of Encoders and Sub-score Ensembling. 
The former technique uses a linear layer to combine  features from different vision encoders, while the latter calculates a weighted sum of sub-scores from independent models for trajectory selection.

%% file: figs/arch.tex
\begin{figure*}[tp]
    \centering
    \includegraphics[width=\linewidth]{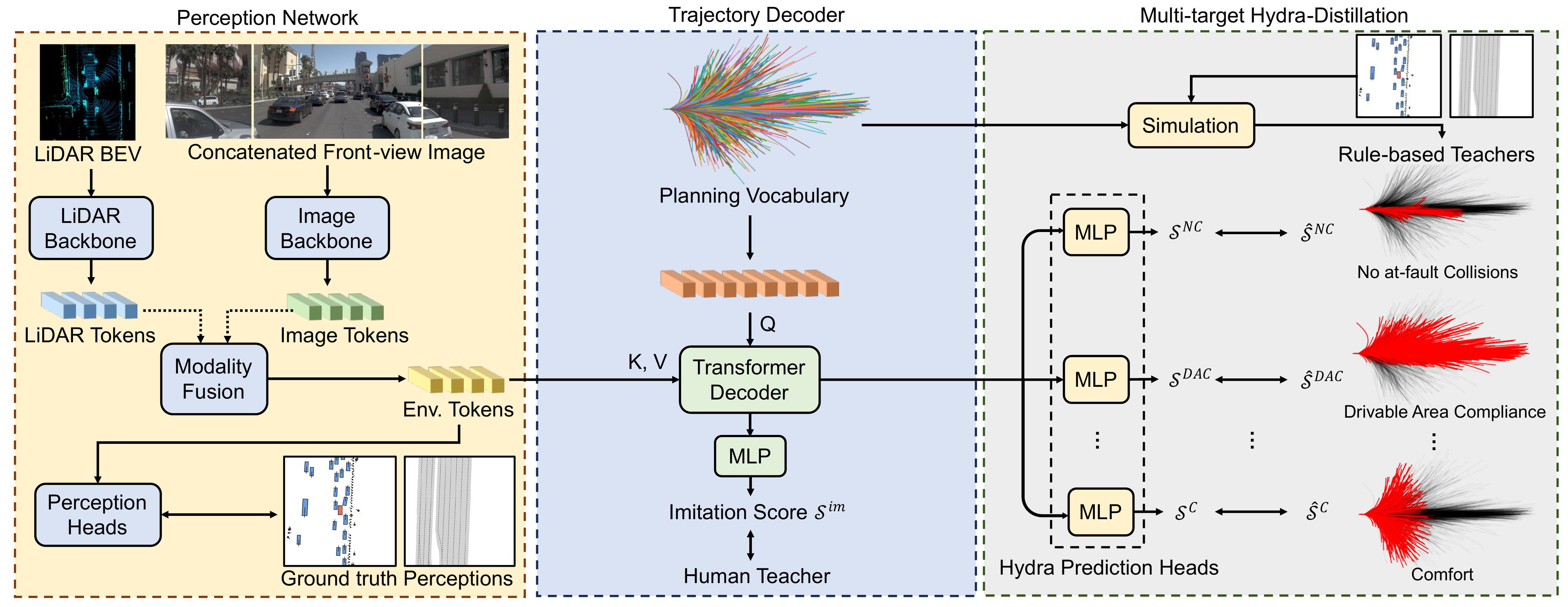}
    \vspace{-0.2in}

    \caption{\textbf{The Overall Architecture of \hydra.}}
    \vspace{-0.2in}

    \label{fig:arch}
\end{figure*}

%% file: 03_experiments.tex
\section{Experiments}
\input{tables/tab1}
\input{tables/tab2}
\subsection{Dataset and metrics}
\vspace{0.05in}\noindent\textbf{Dataset.} The Navsim dataset builds on the existing OpenScene~\cite{openscene2023} dataset, a compact version of nuPlan~\cite{nuplan} with only relevant annotations and sensor data sampled at 2 Hz. The dataset primarily focuses on scenarios involving changes in intention, where the ego vehicle's historical data cannot be extrapolated into a future plan. The dataset provides annotated 2D high-definition maps with semantic categories and 3D bounding boxes for objects.
The dataset is split into two parts: Navtrain and Navtest, which respectively contain 1192 and 136 scenarios for training/validation and testing.

\vspace{0.05in}\noindent\textbf{Metrics.} For this challenge, we evaluate our models based on the PDM score, which can be formulated as follows:
\begin{equation}
    \scalebox{0.8}{$PDM_{score} = NC \times DAC \times DDC \times \frac{(5\times TTC + 2\times C + 5\times EP)}{12},$}
\end{equation}
where sub-metrics $NC$, $DAC$, $TTC$, $C$, $EP$ correspond to the No at-fault Collisions, Drivable Area Compliance, Time to Collision, Comfort, and Ego Progress. 
For the distillation process and subsequent results, $DDC$ is neglected due to an implementation problem.\footnote{https://github.com/autonomousvision/navsim/issues/14}.

\subsection{Implementation Details}
We train our models on the Navtrain split using 8 NVIDIA A100 GPUs, with a total batch size of 256 across 20 epochs. 
The learning rate and weight decay are set to $1\times10^{-4}$ and 0.0 following the official baseline. 
LiDAR points from 4 frames are splatted onto the BEV plane to form a density BEV feature, which is encoded using ResNet34~\cite{he2016deep}.
For images, the front-view image is concatenated with the center-cropped front-left-view and front-right-view images, yielding an input resolution of $256\times1024$ by default. 
ResNet34 is also applied for feature extraction unless otherwise specified. No data or test-time augmentations are used.

\subsection{Main Results}
Our results, presented in Tab.~\ref{table:result}, highlight the absolute advantage of \hydra{} over the baseline. 
In our exploration of different planning vocabularies~\cite{chen2024vadv2}, utilizing a larger vocabulary $\mathcal{V}_{8192}$ demonstrates improvements across different methods. 
Furthermore, non-differentiable post-processing yields fewer performance gains than our framework, while weighted confidence enhances the performance comprehensively.
To ablate the effect of different learning targets, the continuous metric EP (Ego Progress) is not considered in early experiments and we attempt the distillation of the overall PDM score.
Nonetheless, the irregular distribution of the PDM score incurs performance degradation, which suggests the necessity of our multi-target learning paradigm.
In the final version of \hydra{}-$\mathcal{V}_{8192}$-W-EP, the distillation of EP can improve the corresponding metric.

\subsection{Scaling Up and Model Ensembling}
Previous literature~\cite{hu2023planning} suggests larger backbones only lead to minor improvements in planning performance.
Nevertheless, we further demonstrate the scalability of our model with larger backbones.
Tab.~\ref{table:result_scale} shows three best-performing versions of \hydra{} with ViT-L~\cite{yang2024depth, fang2023eva} and V2-99~\cite{lee2019energy} as the image backbone. 
For the final submission, we use the ensembled sub-scores of these three models for inference.

\label{sec:experiment}

%% file: tables/tab1.tex
\begin{table*}[htb]
\scriptsize
\setlength{\tabcolsep}{0.03\linewidth}
\newcommand{\classfreq}[1]{{~\tiny(\semkitfreq{#1}\%)}}  

\def\mystrut{\rule{0pt}{1.5\normalbaselineskip}}
\centering
\begin{tabular}{l| c | c c c c c |c}

    \toprule
    Method 
    & Inputs
    & {NC} 
    & {DAC}
    & {EP} 
    & {TTC} 
    & {C} 
    & {Score}  \\
    \midrule
    
    PDM-Closed~\cite{dauner2023parting}$\diamond$ & Perception GT & 94.6 & 99.8 & 89.9 & 86.9 & 99.9 & 89.1    \\
    \midrule
    Transfuser ~\cite{chitta2022transfuser} & LiDAR \& Camera & 96.5 & 87.9 & 73.9 & 90.2 & 100 & 78.0  \\
    Vadv2-$\mathcal{V}_{4096}$~\cite{chen2024vadv2}* & LiDAR \& Camera & 97.1 & 88.8 & 74.9& 91.4 &100 & 79.7   \\
    Vadv2-$\mathcal{V}_{4096}$~\cite{chen2024vadv2}*-PP & LiDAR \& Camera & 97.0 & 89.1 & 75.0& 91.2 &100 & 79.9   \\
    Vadv2-$\mathcal{V}_{8192}$~\cite{chen2024vadv2}* & LiDAR \& Camera & 97.2 & 89.1 & 76.0& 91.6 &100 & 80.9   \\
    \hydra{}-$\mathcal{V}_{4096}$   & LiDAR \& Camera & 97.7 & 91.5 & 77.5 & 92.7 & 100 & 82.6 \\
    \hydra{}-$\mathcal{V}_{8192}$   & LiDAR \& Camera & 97.9 & 91.7 & 77.6 & 92.9 & 100 & 83.0 \\
    \hydra{}-$\mathcal{V}_{8192}$-PDM   & LiDAR \& Camera & 97.5 & 88.9 & 74.8 & 92.5 & 100 & 80.2 \\
    \hydra{}-$\mathcal{V}_{8192}$-W & LiDAR \& Camera & 98.1 & \textbf{96.1} & 77.8 & 93.9 & 100 & 85.7 \\
    \hydra{}-$\mathcal{V}_{8192}$-W-EP & LiDAR \& Camera & \textbf{98.3} & 96.0 & \textbf{78.7} & \textbf{94.6} & \textbf{100} & \textbf{86.5} \\

\bottomrule
\end{tabular}
\vspace{-2mm}
\caption{\textbf{Performance on the Navtest Split. }
$\diamond$ The official Navsim implementation of PDM-Closed is potentially prone to errors due to inconsistent braking maneuvers and offset formulation compared with the nuPlan implementation~\cite{dauner2023parting}.
All end-to-end methods use the official Transfuser~\cite{chitta2022transfuser} as the perception network. 
* Our distance-based imitation loss is adopted for training.
PP: Transfuser perception is used for post-processing.
PDM: The learning target is the overall PDM score. W: Weighted confidence during inference. EP: The model is trained to fit the continuous EP (Ego Progress) metric.} 
\label{table:result}
\vspace{-0.1in}
\end{table*}

%% file: tables/tab2.tex
\begin{table*}[htb]
\scriptsize
\setlength{\tabcolsep}{0.03\linewidth}
\newcommand{\classfreq}[1]{{~\tiny(\semkitfreq{#1}\%)}}  

\def\mystrut{\rule{0pt}{1.5\normalbaselineskip}}
\centering
\begin{tabular}{l| c c | c c c c c |c}

    \toprule
    Method 
    & Img. Resolution
    & Backbone
    & {NC} 
    & {DAC}
    & {EP} 
    & {TTC} 
    & {C} 
    & {Score}  \\
    \midrule
    
    PDM-Closed~\cite{dauner2023parting}$\diamond$ & - & - & 94.6 & \textbf{99.8} & \textbf{89.9} & 86.9 & 99.9 & 89.1    \\
    \midrule
    \hydra{}-A & $256\times 1024$ & ViT-L* & 98.4 & 97.7 & 85.0 & 94.5 & 100 & 89.9  \\
    \midrule
    \hydra{}-B & $512\times 2048$ & V2-99 & 98.4 & 97.8 & 86.5 & 93.9 & 100 & 90.3  \\
    \midrule
    \multirow{3}{*}{\hydra{}-C} & $256\times 1024$ & ViT-L* & \multirow{3}{*}{\textbf{98.7}} & \multirow{3}{*}{98.2} & \multirow{3}{*}{86.5} & \multirow{3}{*}{\textbf{95.0}} & \multirow{3}{*}{\textbf{100}} & \multirow{3}{*}{\textbf{91.0}}  \\
    & $256\times 1024$ & ViT-L\dag & & & & & & \\ 
    & $512\times 2048$ & V2-99 & & & & & & \\ 

\bottomrule
\end{tabular}
\vspace{-2mm}
\caption{\textbf{The Impact of Scaling Up on the Navtest Split.} 
$\diamond$ The official Navsim implementation of PDM-Closed. * ViT-L is initialized from Depth Anything~\cite{yang2024depth}. \dag ViT-L is EVA~\cite{fang2023eva} pretrained on Objects365~\cite{shao2019objects365} and COCO~\cite{lin2014microsoft}. V2-99~\cite{lee2019energy} is initialized from DD3D~\cite{park2021pseudo}.} 
\vspace{-0.2in}
\label{table:result_scale}
\end{table*}

%% file: 12_appendix.tex
\section{Appendix Section}
\label{sec:appendix_section}
Supplementary material goes here.